\documentclass[11pt]{article}

\usepackage[preprint]{acl}

\usepackage{times}
\usepackage{latexsym}

\usepackage[T1]{fontenc}

\usepackage[utf8]{inputenc}

\usepackage{microtype}

\usepackage{inconsolata}

\usepackage{graphicx}
\usepackage{amsmath}

\usepackage{tikz}
\usetikzlibrary{positioning, calc, backgrounds}
\usepackage{graphicx}
\usepackage{xcolor}
\usepackage{amssymb}
\definecolor{cEgoDark}{gray}{0.18}
\definecolor{cOtherMid}{gray}{0.67}
\definecolor{cPromptBg}{gray}{0.93}
\definecolor{cRatBg}{gray}{0.97}
\definecolor{cBorder}{gray}{0.48}
\title{Responsibility Distribution Estimation in Ego-View Accident Videos\\ with Multimodal Large Language Models}

\author{Ryosei Tamura \\
  Keio University \\
  \texttt{ryoseitamura@keio.jp} \\\And
  Andrew Shin \\
  Keio University \\
  \texttt{shin@inl.ics.keio.ac.jp} \\}

\begin{document}
\maketitle

\begin{abstract}
Recent studies on multimodal traffic accident understanding have mainly relied on infrastructure-camera footage, satellite imagery, or structured crash records. However, such data sources are costly to deploy and maintain at large scale, and they cannot objectively capture what the driver was actually able to observe before the accident. In contrast, ego-view accident videos directly represent the driver's visual perspective, making them suitable for reasoning about avoidability and driver responsibility. In this paper, we introduce \textit{responsibility distribution estimation} for ego-view traffic accident videos, a new task in which a model predicts the percentage of responsibility assigned to each involved agent. We construct an LLM-assisted responsibility annotation pipeline and fine-tune multimodal large language models under multiple input settings, including raw frames, segmentation-enhanced input, and textual descriptions. Experimental results establish a strong initial benchmark, demonstrating that multimodal LLMs can effectively perform this nuanced, constraint-based reasoning task. Our findings suggest that ego-centric accident videos provide a promising foundation for socially and legally meaningful multimodal reasoning beyond conventional accident classification and explanation tasks.
\end{abstract}

\section{Introduction}

Recent advances in large language models (LLMs) and multimodal large language models (MLLMs) have enabled substantial progress in traffic accident analysis, including crash detection, severity prediction, and causal explanation. Existing approaches have primarily relied on structured crash reports, satellite imagery, or infrastructure-camera footage. While these settings are useful for scene-level monitoring, they suffer from two important limitations.

First, infrastructure-based systems are costly to deploy and maintain at scale. Traffic monitoring cameras and high-resolution sensing infrastructure are typically limited to specific intersections or urban regions, making comprehensive coverage difficult. Second, these external viewpoints cannot objectively represent what the driver was actually able to observe before the accident. As a result, they are less suitable for reasoning about avoidability and driver responsibility from the perspective of the vehicle involved in the accident.

In contrast, ego-view accident videos directly capture the driver's visual perspective. This perspective enables driver-centered reasoning, such as whether the ego driver had sufficient time to react, whether another agent suddenly entered the scene, and how responsibility should be distributed among the involved participants. Despite this unique property, prior work on ego-centric accident understanding has mainly focused on accident cause explanation and prevention-oriented reasoning, leaving responsibility allocation largely unexplored.

In this work, we propose \textit{responsibility distribution estimation} for ego-view traffic accident videos. Given an accident video and a set of involved agents, the goal is to predict a responsibility distribution over the agents rather than a single accident label or cause. For example, instead of predicting that a pedestrian alone caused the accident, the model may estimate that the pedestrian is 60\% responsible while the ego driver is 40\% responsible. This formulation allows models to represent shared responsibility and more closely reflects real-world traffic reasoning. To study this problem, we construct an LLM-assisted responsibility labeling pipeline to generate proportional responsibility distributions. We establish the first benchmark for this task by evaluating multimodal models under different input conditions, including image-only input, segmentation-enhanced input, and textual descriptions.

Our work makes three contributions. First, we introduce responsibility distribution estimation as a new task for multimodal accident understanding. Second, we highlight the importance of ego-centric accident videos as a foundation for driver-centered and responsibility-aware reasoning. We establish the first benchmark for this task by evaluating multimodal models under different input conditions.

\section{Related Work}

Recent work has explored the use of LLMs for traffic crash analysis. CrashLLM~\cite{wang2025crashllm} formulates traffic crash records as language and investigates LLM-based prediction and causal analysis for accident outcomes. Other work has used LLMs with chain-of-thought prompting ~\cite{NEURIPS2022_9d560961} and prompt engineering for crash severity prediction~\cite{zhao2024cotseverity}, showing that language models can reason over environmental conditions, driver behavior, and vehicle-related factors. CrashSage~\cite{liu2025crashsage} further proposes an LLM-centered framework for contextual and interpretable crash analysis, combining data verbalization, augmentation, fine-tuning, and attribution-based explanation. These studies demonstrate the potential of LLMs for interpretable accident analysis, but they primarily focus on severity prediction, outcome classification, or causal factor identification rather than proportional responsibility allocation among agents.

Domain-adapted MLLMs have been applied to remote sensing imagery for interpretable road traffic accident analysis~\cite{li2025remote}, where high-resolution overhead images are combined with structured accident data. Other work evaluates MLLMs for accident detection from infrastructure-camera imagery~\cite{chen2025mllmaccident}. These approaches are valuable for accident detection and scene-level explanation, but their visual perspectives differ from egocentric driving videos. 

The MM-AU dataset introduced ego-view accident video understanding, emphasizing accident causes and possible prevention strategies from a driver-centered perspective~\cite{fang2024mmau}. This setting is particularly relevant to responsibility reasoning because the video captures what the ego driver could observe before the accident. Prior use of MM-AU and related datasets has mainly focused on explaining accident causes or generating prevention-oriented descriptions. We extend this direction by asking models to estimate responsibility distributions over involved agents, moving from causal explanation to socially meaningful allocation of responsibility.



CADD~\cite{Shen2025CADDAC} bridges behavioral understanding in dashcam videos with statute-based liability attribution by utilizing a framework to map driving actions directly to specific traffic laws for definitive liability judgments. While their work emphasizes strict legal and statutory grounding for liability, our approach focuses on estimating the continuous proportional responsibility distribution among involved agents using generative multimodal reasoning.

Compared with prior work on crash severity prediction, accident detection, and natural-language causal explanation, our work is significantly different in following ways. First, we focus on egocentric accident videos rather than structured crash records, satellite imagery, or infrastructure-camera images. Second, we predict a responsibility distribution rather than a single accident type, severity label, or textual cause. Although our current labels are LLM-assisted and should not be treated as legal ground truth, the task provides a practical framework for studying whether multimodal models can produce structured, constraint-satisfying responsibility judgments from accident videos.


\section{Method}

\subsection{Task Definition}

We propose \textit{Responsibility Distribution Estimation} for ego-view traffic accident videos. Given an accident video $V$ and a set of involved agents $A = \{a_1, a_2, \dots, a_n\}$, the goal is to predict a responsibility distribution $\mathbf{r} = [r_1, r_2, \dots, r_n]$, where each $r_i \in [0,100]$ represents the percentage of responsibility assigned to agent $a_i$, subject to $\sum_{i=1}^{n} r_i = 100$.

Unlike conventional accident classification tasks, our objective is not to identify a single cause or responsible agent, but rather to estimate how responsibility is distributed among multiple participants. In our setting, the agents may include the ego vehicle, other vehicles, pedestrians, bicycles, or motorcycles. Because the videos are ego-centric, the task additionally requires reasoning about what information was available to the ego driver before the accident. We also require the model to identify which party is primarily responsible and to label the type of the other party.

\subsection{Dataset Construction}

We build our dataset on top of MM-AU~\cite{fang2024mmau}, which provides ego-view dashcam accident videos with text annotations describing the accident scene, its causes, and suggested preventive measures. Since MM-AU does not include responsibility labels, we construct an automated annotation pipeline with LLM as an annotator.

For each accident video, we uniformly sample $k=8$ frames spanning the annotated abnormal event window, ensuring the accident frame is always included. We present these frames to LLM alongside the scene metadata and all available text annotations. The model is prompted to output (i) the integer percentages of responsibility assigned to the ego vehicle and other parties, (ii) which party is the primary responsible party, (iii) the category of the other parties (\texttt{pedestrian}, \texttt{other\_vehicle}, \texttt{motorcycle\_cyclist}, \texttt{environment}, or \texttt{unknown}), and (iv) a rationale grounding the judgment in the visual and textual evidence. We use temperature zero to maximize label consistency, and ensure that percentages sum to exactly 100. The resulting dataset is roughly balanced between ego-primary (52\%) and other-party-primary (48\%) cases. The most frequent other-party types are other vehicles (58\%) and motorcycle cyclists (33\%), with pedestrians accounting for 7\%. 

\subsection{Model and Training}

We fine-tune Qwen3-VL-4B~\cite{qwen3vl} on the annotated responsibility dataset with LoRA adaptation~\cite{hu2022lora} to all linear projection layers with rank $r=16$, scaling factor $\alpha=32$, and dropout 0.05. Given a multimodal input $X$ consisting of $k$ accident frames and an instruction prompt, the model autoregressively generates a JSON-formatted output containing the predicted responsibility percentages, the primary responsible party, the other-party label, and a brief rationale. 

Each training example follows an instruction-tuning format. The user turn contains the case metadata, text annotations, and $k=10$ sampled frames. The assistant turn contains the target output derived from the LLM annotations. Distribution validity is enforced implicitly: the prompt instructs the model that the percentages must sum up to 100, and all training data satisfy this constraint, so the model learns the requirement from the data.


\section{Experiment}

\begin{figure*}[t]
\centering
\resizebox{\textwidth}{!}{%
\begin{tikzpicture}[
  inner sep = 0pt,
  outer sep = 0pt,
  font      = \scriptsize,
]

\foreach \i/\fname in {
  1/frame_0100, 2/frame_0116, 3/frame_0131,
  4/frame_0147, 5/frame_0162, 6/frame_0178,
  7/frame_0193, 8/frame_0209, 9/frame_0224,
  10/frame_0240}
{
  \pgfmathsetmacro{\xnw}{(\i-1)*1.65}
  \node[draw=cBorder, line width=0.25pt, anchor=north west]
    (fr\i) at (\xnw cm, 0cm) {%
      \includegraphics[width=1.53cm, height=1.15cm,
                       keepaspectratio=false]{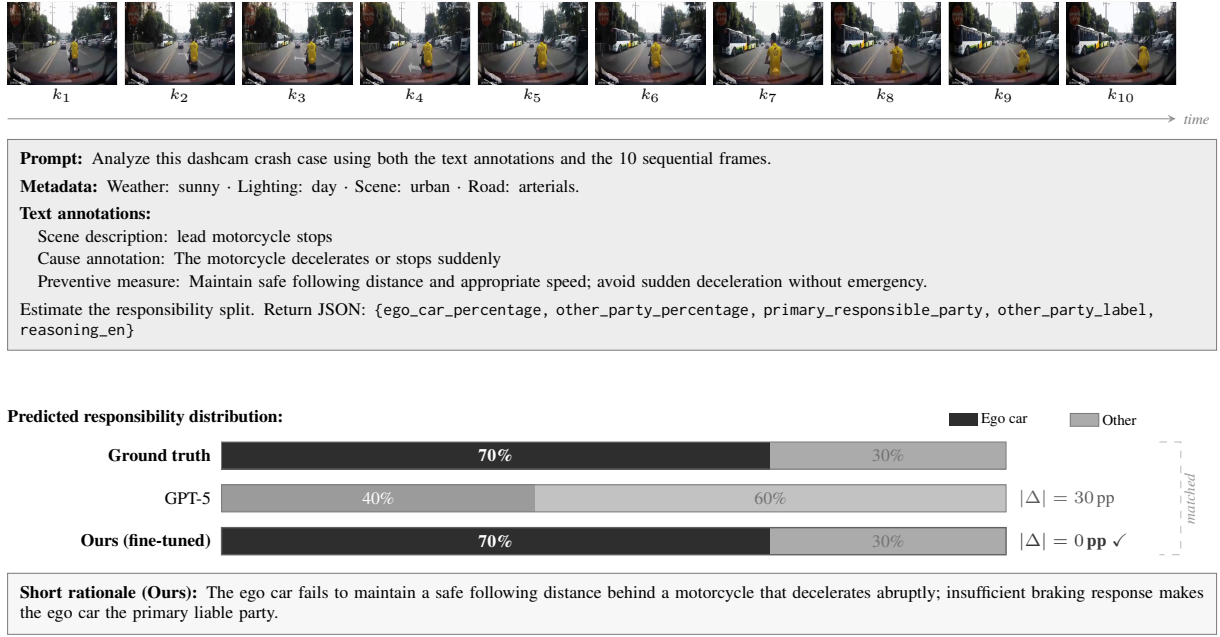}%
  };
  \node[anchor=north, font=\tiny, inner sep=1.5pt]
    at (fr\i.south) {\itshape$k_{\i}$};
}

\draw[->, >=stealth, line width=0.45pt, color=black!40]
  (0cm, -1.63cm) -- (16.38cm, -1.63cm);
\node[anchor=west, color=black!50, font=\tiny\itshape, inner sep=2pt]
  at (16.42cm, -1.63cm) {time};

\node[
  fill=cPromptBg, draw=cBorder, line width=0.3pt,
  anchor=north west,
  text width=16.60cm,
  align=left,
  inner sep=5pt,
] (pbox) at (0cm, -1.92cm) {%
  \textbf{Prompt:}
  Analyze this dashcam crash case using both the text annotations
  and the 10 sequential frames.\\[3pt]
  \textbf{Metadata:}\enspace
  Weather: sunny~$\cdot$~Lighting: day~$\cdot$~Scene: urban~$\cdot$~Road: arterials.\\[3pt]
  \textbf{Text annotations:}\\[1pt]
  \quad Scene description:\enspace lead motorcycle stops\\[1pt]
  \quad Cause annotation:\enspace The motorcycle decelerates or stops suddenly\\[1pt]
  \quad Preventive measure:\enspace Maintain safe following distance and appropriate
  speed; avoid sudden deceleration without emergency.\\[3pt]
  Estimate the responsibility split. Return JSON:\enspace
  \texttt{\{ego\_car\_percentage, other\_party\_percentage,
  primary\_responsible\_party, other\_party\_label, reasoning\_en\}}%
};


\node[anchor=west, font=\scriptsize\bfseries]
  at (0cm, -5.82cm) {Predicted responsibility distribution:};

\fill[cEgoDark]  (13.20cm, -5.94cm) rectangle (13.60cm, -5.76cm);
\fill[cOtherMid] (14.90cm, -5.94cm) rectangle (15.30cm, -5.76cm);
\draw[cBorder, line width=0.2pt]
  (14.90cm, -5.94cm) rectangle (15.30cm, -5.76cm);
\node[anchor=west, font=\tiny] at (13.65cm, -5.85cm) {Ego car};
\node[anchor=west, font=\tiny] at (15.35cm, -5.85cm) {Other};

\node[anchor=east, font=\scriptsize\bfseries, inner sep=2pt]
  at (2.92cm, -6.35cm) {Ground truth};
\fill[cEgoDark]  (3.00cm, -6.54cm) rectangle (10.70cm, -6.16cm);
\fill[cOtherMid] (10.70cm, -6.54cm) rectangle (14.00cm, -6.16cm);
\draw[cBorder, line width=0.25pt]
  (3.00cm, -6.54cm) rectangle (14.00cm, -6.16cm);
\node[white, font=\scriptsize\bfseries] at (6.85cm,  -6.35cm) {70\%};
\node[black!55, font=\scriptsize]       at (12.35cm, -6.35cm) {30\%};

\node[anchor=east, font=\scriptsize, inner sep=2pt]
  at (2.92cm, -6.95cm) {GPT-5};
\fill[cEgoDark!48]  (3.00cm, -7.14cm) rectangle (7.40cm,  -6.76cm);
\fill[cOtherMid!72] (7.40cm, -7.14cm) rectangle (14.00cm, -6.76cm);
\draw[cBorder, line width=0.25pt]
  (3.00cm, -7.14cm) rectangle (14.00cm, -6.76cm);
\node[white, font=\scriptsize]    at (5.20cm,  -6.95cm) {40\%};
\node[black!55, font=\scriptsize] at (10.70cm, -6.95cm) {60\%};
\node[anchor=west, font=\scriptsize, color=black!65]
  at (14.18cm, -6.95cm) {$|\Delta|=30$\,pp};

\node[anchor=east, font=\scriptsize\bfseries, inner sep=2pt]
  at (2.92cm, -7.55cm) {Ours (fine-tuned)};
\fill[cEgoDark]  (3.00cm, -7.74cm) rectangle (10.70cm, -7.36cm);
\fill[cOtherMid] (10.70cm, -7.74cm) rectangle (14.00cm, -7.36cm);
\draw[black!62, line width=0.55pt]
  (3.00cm, -7.74cm) rectangle (14.00cm, -7.36cm);
\node[white, font=\scriptsize\bfseries] at (6.85cm,  -7.55cm) {70\%};
\node[black!55, font=\scriptsize]       at (12.35cm, -7.55cm) {30\%};
\node[anchor=west, font=\scriptsize\bfseries, color=black!78]
  at (14.18cm, -7.55cm) {$|\Delta|=0$\,pp\;$\checkmark$};

\draw[black!28, line width=0.35pt, dashed]
  (16.15cm, -6.16cm) -- (16.45cm, -6.16cm)
  -- (16.45cm, -7.74cm)
  -- (16.15cm, -7.74cm);
\node[rotate=90, anchor=south, font=\tiny\itshape, color=black!40]
  at (16.65cm, -6.95cm) {matched};

\node[
  fill=cRatBg, draw=cBorder, line width=0.25pt,
  anchor=north west,
  text width=16.60cm,
  align=left,
  inner sep=5pt,
] at (0cm, -8.00cm) {%
  \textbf{Short rationale (Ours):}\enspace
  The ego car fails to maintain a safe following distance behind a
  motorcycle that decelerates abruptly; insufficient braking response
  makes the ego car the primary liable party.%
};

\end{tikzpicture}%
}
\caption{Qualitative input--output comparison on a held-out test
  case from MM-AU.
  \textbf{Top}: ten uniformly sampled ego-view frames spanning
  the accident window ($k_1$--$k_{10}$).
  \textbf{Middle}: the actual prompt given to both models, including
  scene metadata and text annotations.
  \textbf{Bottom}: predicted responsibility distributions
  (\textit{dark} = ego car; \textit{light} = other party) for
  GPT-5 and our fine-tuned model, compared with the ground truth.
  GPT-5 misidentifies the primary responsible party and incurs a
  30-percentage-point error; our model matches the ground truth
  exactly ($|\Delta|=0$\,pp).}
\label{fig:qualitative}
\vspace{-3mm}
\end{figure*}

\subsection{Setting}

\textbf{Input:} To analyze which sources of information contribute to responsibility estimation, we evaluate three input conditions. \textbf{RGB} provides the model with $k=10$ uniformly sampled accident frames together with all text annotations. \textbf{RGB+Seg} replaces the RGB frames with semantic segmentation overlays extracted from MM-AU while keeping the same text annotations, allowing us to test whether structured visual representations are more informative than raw appearance. \textbf{Text-only} provides only the scene description and no images, isolating the contribution of visual input.

\textbf{Baseline:} We compare our fine-tuned Qwen3-VL-4B against GPT-5 used in a zero-shot setting under the same three input conditions. For each condition, the same structured JSON prompt is used for both models, ensuring a fair comparison. The zero-shot baseline assesses the out-of-the-box capability of a strong frontier model on our task.








\textbf{Evaluation metrics:} We report the following metrics. \textbf{Exact Match} (\%): the proportion of cases where the predicted ego-vehicle responsibility percentage equals the ground-truth label exactly. \textbf{Within-10\%} and \textbf{Within-20\%}: the proportion of cases where the absolute error is at most 10 or 20 percentage points, measuring practical tolerance. \textbf{MAE} and \textbf{RMSE}: mean absolute error and root mean squared error of the ego-vehicle responsibility percentage. \textbf{Primary Party Acc.} (\%): accuracy of predicting which party bears primary responsibility, a classification sub-task within the distribution estimation problem.

\subsection{Results and Discussion}

\begin{table}[t]
\centering
\renewcommand{\arraystretch}{1.15}
\resizebox{\columnwidth}{!}{%
\begin{tabular}{llccccc}
\hline
\textbf{Model} & \textbf{Input} & \textbf{EM} & \textbf{W-10} & \textbf{W-20} & \textbf{MAE} & \textbf{PP} \\
\hline
GPT-5 (ZS)   & RGB      & 38.5 & 79.1 & 86.8 & 12.1 & 89.0 \\
GPT-5 (ZS)   & RGB+Seg  & 28.6 & 79.1 & 87.9 & 12.7 & 89.0 \\
GPT-5 (ZS)   & Text     & 23.1 & 45.1 & 57.1 & 24.9 & 62.6 \\
\hline
Qwen (FT)    & RGB      & \textbf{79.1} & \textbf{90.1} & \textbf{95.6} & \textbf{4.9} & \textbf{96.7} \\
Qwen (FT)    & RGB+Seg  & \textbf{79.1} & \textbf{90.1} & 94.5 & 5.5 & 95.6 \\
Qwen (FT)    & Text     & 24.2 & 51.6 & 57.1 & 35.1 & 59.3 \\
\hline
\end{tabular}%
}
\caption{Test-set results. EM: Exact Match; W-10/W-20: within 10/20 percentage points; MAE: mean absolute error on ego-vehicle percentage; PP: Primary Party accuracy. ZS: zero-shot; FT: LoRA fine-tuned.}
\label{tab:main}
\vspace{-3mm}
\end{table}

Table~\ref{tab:main} presents the main results. Our fine-tuned Qwen3-VL-4B demonstrates that models can successfully learn to allocate proportional responsibility, successfully adapting to the constraints of the task compared to frontier models. Under RGB input, the fine-tuned model achieves an Exact Match of 79.1\% and a Primary Party accuracy of 96.7\%, compared to 38.5\% and 89.0\% for GPT-5 zero-shot. MAE drops from 12.1 to 4.9 percentage points, a reduction of more than half.

As expected, both the fine-tuned model and GPT-5 degrade sharply when visual input is removed. Under text-only input, GPT-5 MAE more than doubles (12.1 $\to$ 24.9) and Primary Party accuracy drops to 62.6\%. The fine-tuned model degrades even more severely on MAE (4.9 $\to$ 35.1). These results confirm that ego-view video frames carry information about responsibility that cannot be recovered from textual descriptions alone.

Figure~\ref{fig:qualitative} shows a qualitative input-output example. Both GPT-5 and our fine-tuned model receive the same ego-view frames, metadata, text annotations, and structured prompt. While GPT-5 reverses the primary responsible party, our model predicts the same responsibility distribution as the ground-truth label, demonstrating that training with LLM-annotated allocation helps the model better map multimodal accident evidence to the
responsibility distribution. Semantic segmentation overlays yield comparable or slightly worse results for both models, with Primary Party accuracy dropping by 1--2 points and MAE increasing slightly, suggesting that the segmentation representation hardly adds information beyond what RGB frames. Structured visual representations combined with more explicit reasoning scaffolds may be beneficial.

GPT-5 zero-shot already achieves high Within-10\% (79.1\%) under RGB input, indicating that the zero-shot model can reason roughly correctly about which direction responsibility falls. However, our fine-tuned model's Exact Match of 79.1\% shows that fine-tuning on LLM-annotated labels substantially anchors the model to the specific percentage values present in the training distribution, rather than merely ranking parties correctly. 


\section{Conclusion \& Future Work}
We proposed a novel responsibility distribution estimation task for ego-view traffic accident videos to assign proportional responsibility to the involved agents. We constructed an LLM-assisted annotation and training pipeline, and our experiments demonstrate that multimodal models can successfully navigate the nuanced reasoning required for proportional responsibility allocation. Future research should further validate the LLM-assisted labels with human experts, expand the dataset to more diverse traffic scenarios, and incorporate legal context into responsibility estimation. 

\section*{Limitations}

This study has several limitations. First, the responsibility labels are generated with LLM assistance and should not be treated as human expert annotations or legal ground truth. The current annotation process does not explicitly account for jurisdiction-specific legal context, which limits its ability to distinguish causal contribution from legal responsibility.

Second, our model relies on sampled frames and available text annotations rather than explicit physical evidence such as speed, braking behavior, agent trajectories, or reaction time. The model may therefore miss fine-grained temporal cues that are important for responsibility estimation.

Third, our current formulation mainly considers responsibility allocation between the ego vehicle and another involved party. Accidents involving three or more interacting agents may require more complex responsibility decomposition, since the model must distinguish each agent's contribution and their interactions before the collision.

\section*{Ethical Considerations}
The estimation of responsibility in traffic accidents is a highly sensitive task with significant legal, financial, and personal implications. While our proposed framework demonstrates the potential for driver-centered multimodal accident reasoning, it is crucial to emphasize that this system is an exploratory research prototype and must not be used for automated legal, judicial, or insurance decision-making. Primarily, our training dataset utilizes LLM-assisted responsibility labels; these serve as a proxy to evaluate multimodal constraint-satisfaction and reasoning, but they are not verified legal ground truths or human expert judgments. Furthermore, the model evaluates scenarios based purely on visual and text-annotated sequences, lacking the capacity to account for jurisdiction-specific statutory traffic laws or fine-grained physical telemetry, such as exact vehicle speed and braking force. Relying strictly on such a model in real-world settings risks automation bias, where a system's misinterpretation of an ego-view perspective could lead to an unjust distribution of fault. Therefore, any future real-world application of this technology necessitates a strict human-in-the-loop architecture, wherein the model functions solely as an assistive tool to help investigators triage multimodal evidence, rather than replacing the legally binding and nuanced judgments of traffic authorities. Finally, because dashcam videos inherently capture personally identifiable information such as license plates and faces, researchers expanding on this dataset must employ rigorous anonymization protocols to protect the privacy of all recorded agents.

\bibliography{custom}

@article{fang2024mmau,
  title={Abductive Ego-View Accident Video Understanding for Safe Driving Perception},
  author={Jianwu Fang and Lei-lei Li and Junfei Zhou and Junbin Xiao and Hongkai Yu and Chen Lv and Jianru Xue and Tat-Seng Chua},
  journal={2024 IEEE/CVF Conference on Computer Vision and Pattern Recognition (CVPR)},
  year={2024},
  pages={22030-22040},
  url={https://api.semanticscholar.org/CorpusID:268201615}
}

@article{wang2025crashllm,
  title={Learning Traffic Crashes as Language: Datasets, Benchmarks, and What-if Causal Analyses},
  author={Zhiwen Fan and Pu Wang and Yang Zhao and Yibo Zhao and B. Ivanovic and Zhangyang Wang and Marco Pavone and Hao Frank Yang},
  journal={ArXiv},
  year={2024},
  volume={abs/2406.10789},
  url={https://api.semanticscholar.org/CorpusID:270560086}
}

@article{li2025remote,
  title={Domain-Adapted MLLMs for Interpretable Road Traffic Accident Analysis Using Remote Sensing Imagery},
  author={Bing He and Wei He and Qing Chang and Wen Luo and Ling Xiao},
  journal={ISPRS Int. J. Geo Inf.},
  year={2025},
  volume={15},
  pages={8},
  url={https://api.semanticscholar.org/CorpusID:284211543}
}

@article{chen2025mllmaccident,
  title={Investigating Traffic Accident Detection by Using Multimodal Large Language Models},
  author={Ilhan Skender and Kailin Tong and Selim Solmaz and Daniel Watzenig},
  journal={2025 IEEE International Automated Vehicle Validation Conference (IAVVC)},
  year={2025},
  pages={1-7},
  url={https://api.semanticscholar.org/CorpusID:281495787}
}

@article{zhao2024cotseverity,
  title={Leveraging Large Language Models with Chain-of-Thought and Prompt Engineering for Traffic Crash Severity Analysis and Inference},
  author={Hao Zhen and Yucheng Shi and Yongcan Huang and Jidong J. Yang and Ninghao Liu},
  journal={ArXiv},
  year={2024},
  volume={abs/2408.04652},
  url={https://api.semanticscholar.org/CorpusID:271843060}
}

@article{liu2025crashsage,
  title={CrashSage: A Large Language Model-Centered Framework for Contextual and Interpretable Traffic Crash Analysis},
  author={Hao Zhen and Jidong J. Yang},
  journal={ArXiv},
  year={2025},
  volume={abs/2505.07853},
  url={https://api.semanticscholar.org/CorpusID:278534679}
}

@inproceedings{NEURIPS2022_9d560961,
 author = {Wei, Jason and Wang, Xuezhi and Schuurmans, Dale and Bosma, Maarten and ichter, brian and Xia, Fei and Chi, Ed and Le, Quoc V and Zhou, Denny},
 booktitle = {Advances in Neural Information Processing Systems},
 editor = {S. Koyejo and S. Mohamed and A. Agarwal and D. Belgrave and K. Cho and A. Oh},
 pages = {24824--24837},
 publisher = {Curran Associates, Inc.},
 title = {Chain-of-Thought Prompting Elicits Reasoning in Large Language Models},
 url = {https://proceedings.neurips.cc/paper_files/paper/2022/file/9d5609613524ecf4f15af0f7b31abca4-Paper-Conference.pdf},
 volume = {35},
 year = {2022}
}

@article{qwen3vl,
  title={Qwen3 Technical Report},
  author={Qwen Team},
  journal={arXiv preprint arXiv:2505.09388},
  year={2025}
}

@inproceedings{hu2022lora,
  title={LoRA: Low-Rank Adaptation of Large Language Models},
  author={Hu, Edward J and Shen, Yelong and Wallis, Phillip and Allen-Zhu, Zeyuan and Li, Yuanzhi and Wang, Shean and Wang, Lu and Chen, Weizhu},
  booktitle={International Conference on Learning Representations},
  year={2022}
}

@article{Shen2025CADDAC,
  title={CADD: A Chinese Traffic Accident Dataset for Statute-Based Liability Attribution},
  author={Yunfei Shen and Zhongcheng Wu},
  journal={ArXiv},
  year={2025},
  volume={abs/2511.11715},
  url={https://api.semanticscholar.org/CorpusID:283072079}
}




\end{document}